# TURNABOUTLLM: A Deductive Reasoning Benchmark from Detective Games

**Yuan Yuan** * **Muyu He** * **Muhammad Adil Shahid**
**Jiani Huang** **Ziyang Li** **Li Zhang**
University of Pennsylvania Drexel University
{yyuan86|muyuhe}@upenn.edu {harry.zhang}@drexel.edu

## Abstract

This paper introduces TURNABOUTLLM , a novel framework and dataset for evaluating the deductive reasoning abilities of Large Language Models (LLMs) by leveraging the interactive gameplay of detective games Ace Attorney and Danganronpa. The framework tasks LLMs with identifying contradictions between testimonies and evidences within long narrative contexts, a challenging task due to the large answer space and diverse reasoning types presented by its questions. We evaluate twelve state-of-the-art LLMs on the dataset, hinting at limitations of popular strategies for enhancing deductive reasoning such as extensive thinking and Chain-of-Thought prompting. The results also suggest varying effects of context size, the number of reasoning step and answer space size on model performance. Overall, TURNABOUTLLM presents a substantial challenge for LLMs' deductive reasoning abilities in complex, narrative-rich environments.[1]

## 1 Introduction

Detective stories contain some of the most difficult reasoning problems, meticulously crafted to be intriguing and illusive for even the most intelligent readers. To perform said deduction requires various abilities. Some include information retrieval from long passages of narrative with attention to particular details. Others include piecing together facts with knowledge of physical laws, social norms, timeline of events, and so on. As large language models (LLMs) are increasingly coveted for their reasoning ability, evaluating them on detective stories brings about unique challenges.

Unfortunately, evaluating LLMs' deductive reasoning via detective stories is often infeasible. For example, Sherlock Holmes involves rich reasoning but does not contain explicit questions to pose to models. As a result, existing work that leveraged detective stories for evaluation either only considered simple snippets as the context (Del and Fishel, 2023a) or character relationship prediction as the task (Zhao et al., 2024). Some also focus on textual understandings that require simple reasoning abilities (Xu et al., 2025). To overcome this limitation, we take advantage of a unique asset, detective games, as their interactive gameplay provides a natural interface for evaluating LLMs.

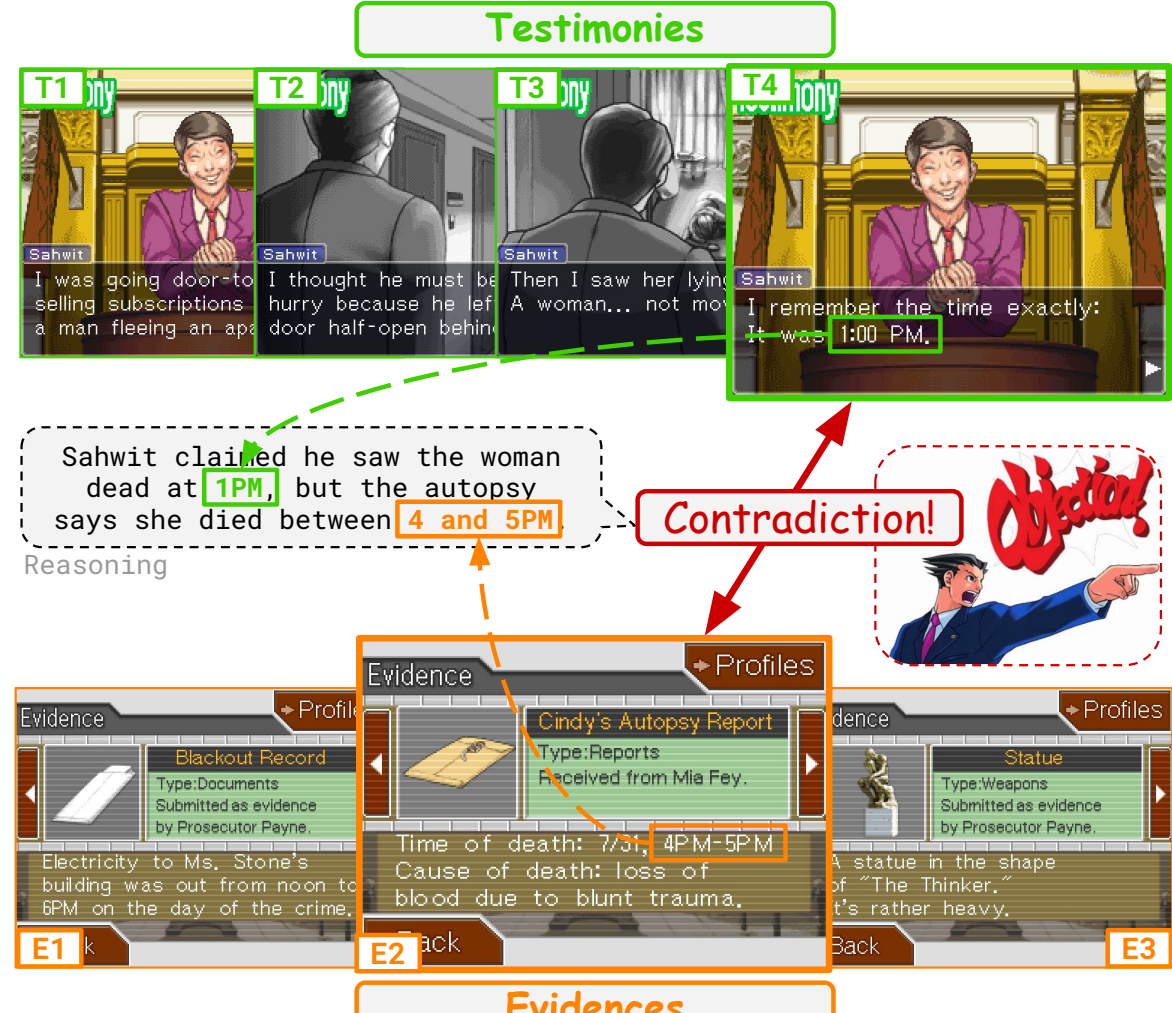

Figure 1: An illustration of a problem from Ace Attorney, a detective game where players are instructed to pinpoint a contradiction between a piece of evidence and a testimony. Adapted to a task in TURNABOUTLLM, the input is a list of testimonies and a list of evidences with their corresponding textual descriptions. The output is the pair of testimony (T4) and evidence (E2) that contradict each other. The example shown is from the introductory episode and is likely the easiest.

We propose TURNABOUTLLM[2], a framework and textual dataset to evaluate LLMs' deductive reasoning ability in a long narrative context. TURN-

* Equal contribution.

[1] Our resources can be found at https://github.com/zharry29/turnabout_llm.

[2] The name "Turnabout" is a wordplay from Ace Attorney as a nod to the playable character's knack for completely changing the direction of a trial, against all odds.

| **Dataset** | **Sym.** | **SLC** | **LAS** | **Nat.** | **MH** | **Het.** |
|---|---|---|---|---|---|---|
| BIG-Bench Hard | ✗ | ✗ | ✗ | ✓ | ✓ | ✗ |
| LogicQA | ✗ | ✗ | ✗ | ✓ | ✓ | ✗ |
| ReClor | ✗ | ✗ | ✗ | ✓ | ✓ | ✗ |
| ZebraLogic | ✗ | ✗ | ✓ | ✓ | ✓ | ✗ |
| ProofWriter | ✓ | ✗ | ✗ | ✗ | ✓ | ✗ |
| FOLIO | ✓ | ✗ | ✗ | ✓ | ✓ | ✗ |
| ProntoQA | ✓ | ✗ | ✗ | ✗ | ✗ | ✗ |
| LogicBench | ✓ | ✗ | ✗ | ✗ | ✗ | ✗ |
| *TurnaboutLLM* | ✓ | ✓ | ✓ | ✓ | ✓ | ✓ |

Table 1: Qualitative comparison of TURNABOUTLLM against other deductive reasoning benchmarks. There are no previous benchmarks that satisfy all six desiderata simultaneously. Our proposed TURNABOUTLLM is the first benchmark to include *symbolic logical annotations* (Sym.) for reasoning tasks situated in *natural scenarios* (Nat.) with *super-long contexts* (SLC), *large answer spaces* (LAS), *multi-hop* (MH) reasoning steps, and *heterogeneous* (Het.) reasoning types.

ABOUTLLM is constructed using two critically acclaimed detective games Ace Attorney[3] and Danganronpa[4]. The core gameplay mechanism, adapted as our task format, is to read through a story, examine existing evidences, examine witness testimonies, deduce likely conclusions, and find a contradiction between an evidence and a testimony in each turn of gameplay, all in text. One example from the 306 turns can be seen in Figure 1. TURNABOUTLLM is superior to existing reasoning benchmarks in that:

1. it includes natural contexts written by human authors that sometimes exceeds 100K words;
2. it presents a large answer space that can contain 300 candidate answers;
3. it contains rigorous yet *heterogeneous* questions that demands temporal, spatial, behavior, object state, causal and numerical understanding;
4. all of the examples contain expert annotations of evidence spans, context summary, reasoning type, and the complete reasoning steps.

We conducted 26 experiments on 12 state-of-the-art LLMs using TURNABOUTLLM, revealing several intriguing insights detailed in Section 5. The results establish TURNABOUTLLM as a substantial challenge for current LLMs outside their training corpus, as the top-performing DeepSeek-R1 only obtains an accuracy score of 45.72%. We observe the generation of extensive reasoning tokens does not directly help with model performance but is negatively correlated with accuracy. The traditionally effective Chain-of-Thought prompting method also presents minimal benefits on complex deductive tasks. When presented with excessive contextual information, only large models, not small and medium-sized ones, can leverage needle-in-a-haystack retrieval to improve reasoning outcomes. We find that performance declines as the number of reasoning steps increases but is unaffected by the size of the answer space, and conversely performance improves with larger parameter counts.

## 2 Related Work

**General Reasoning Benchmarks** To broadly assess models' reasoning capacities, multiple general-purpose benchmarks have been widely studies. They include MMLU (Hendrycks et al., 2021), SuperGLUE (Wang et al., 2020), BIG-Bench (Srivastava et al., 2023), and BIG-Bench Hard (Suzgun et al., 2022). While these benchmarks provide a useful overview, they are not exclusively focused on reasoning tasks, resulting in a limited reflection of models' actual reasoning skills.

In contrast, several benchmarks explicitly target deductive reasoning capacities. LogiGLUE (Luo et al., 2024) integrates 24 reasoning-focused datasets into a unified benchmark. LogiQA (Liu et al., 2020) and ReClor (Yu et al., 2020) draw logical reasoning questions from standardized exams like the LSAT in multi-choice formats. ZebraLogic (Lin et al., 2025) constructs constraint-satisfaction problems that feature expansive answer spaces. However, these benchmarks lack symbolic annotations of logical structures, limiting insights into underlying reasoning processes.

**Synthetic Datasets for LLM Reasoning** Synthetic datasets fulfill the need for symbolic annotations by using LLMs to generate examples based on logical rules. PrOntoQA (Saparov and He, 2023) and LogicBench (Parmar et al., 2024) synthesize questions from logical rules applied to ontological entities, while JustLogic (Chen et al., 2025) uses randomly sampled real-world sentences as premises for reasoning chains. Nonetheless, they typically focus on single inference rules rather than multi-hop reasoning. To address this gap, Multi-LogiEval (Patel et al., 2024) and ProofWriter (Tafjord et al., 2021), an improvement to RuleTaker (Clark et al., 2020), require models to validate synthetic conclusions involving multiple logical steps. However, along with the expert-curated multi-hop FOLIO (Han et al., 2024), these datasets suffer from limited context sizes and answer spaces.

[3]https://en.wikipedia.org/wiki/Ace_Attorney
[4]https://en.wikipedia.org/wiki/Danganronpa

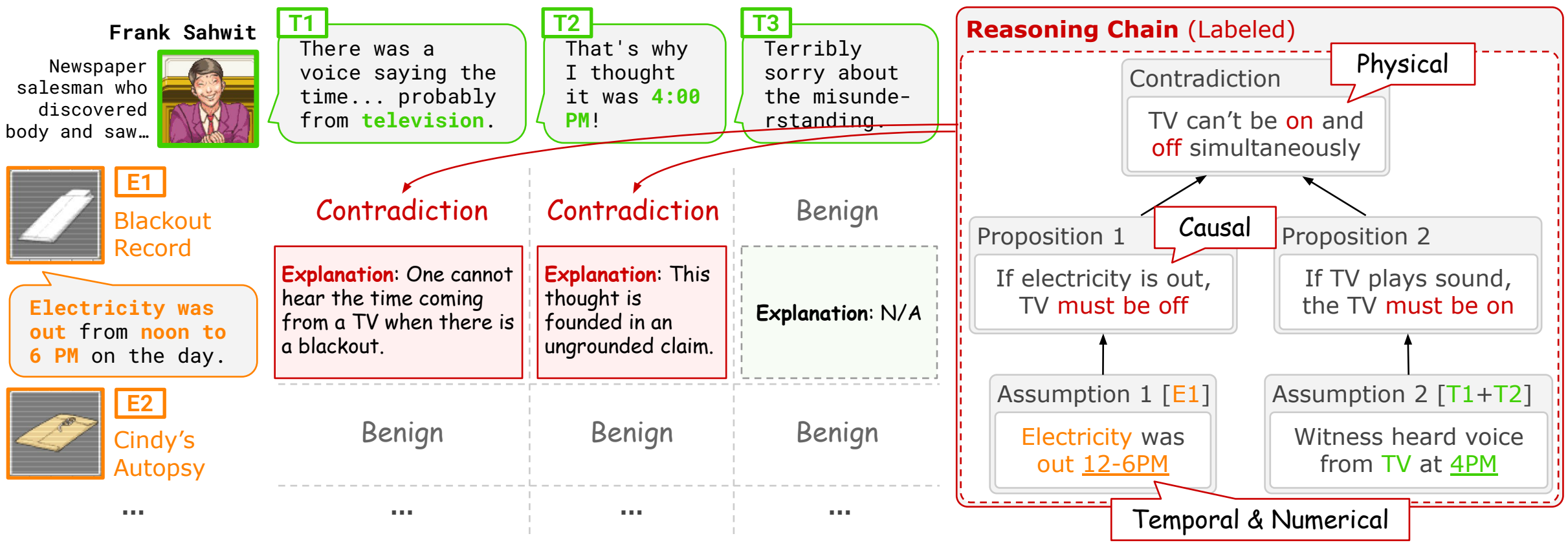

Figure 2: An example data point from TURNABOUTLLM, where testimonies, marked as T1 to T3, are shown horizontally in green and evidences E1, E2 and more are shown vertically in orange. In addition to labeling which testimony-evidence pairs are contradictory, we provide a per-contradiction explanation and a ground-truth reasoning chain used to derive the contradiction. Each reasoning chain forms a tree structure: leaf nodes represent observed facts, while internal (non-leaf) nodes correspond to intermediate atomic propositions that perform derivations.

**Reasoning Datasets from Detective Stories** Detective stories naturally engage readers in multi-hop deduction, thus well-suited for deductive reasoning evaluations. MuSR (Sprague et al., 2024) and True Detective (Del and Fishel, 2023b) synthesize detective stories from predefined facts or online detective games, yet they face inherent limitations of small context sizes. Benchmarks derived from authentic novels or high-quality puzzles, such as WhoDunIt (Gupta, 2025), DetectBench (Gu et al., 2024), and DetectiveQA (Xu et al., 2025), address this context size limitation. However, their answer spaces remain relatively constrained. To the best of our knowledge, there is no existing benchmark that leverages the detective story format to combine symbolic annotations with reasoning tasks characterized by large contexts and answer spaces. A comprehensive overview of each benchmark's attributes is presented in Table 1.

## 3 Dataset and Task

Our TURNABOUTLLM dataset is based on 11 titles of critically acclaimed Ace Attorney series and Danganronpa. In this section, we detail our process of creating the TURNABOUTLLM dataset (Section 3.1), the additional annotations (Section 3.2), and the overall statistics (Section 3.3).

### 3.1 Data Creation

**Extraction** To obtain data, we crawl and parse an Ace Attorney Wiki[5] and a Danganronpa archive[6]. We extract the following data: 1) **character information**, including name, gender, age, and a description; 2) **evidence information**[7], including name, source, and a description; 3) **testimonies** in the core gameplay[8], including speaker, content, and the correct evidence to present if the testimony can be contradicted; and 4) **transcript** of the full gameplay[9], including dialogues, information, and flavor text, used as the full context. While the games are originally visual novels in nature, we only consider the textual elements, which are sufficient for reasoning in most cases. Whenever visuals are indispensable for reasoning, they are manually captioned so that key visual features are provided.

**Modification** Using the data acquired above, we construct each each example, referred to as a turn, as follows. The input to a model is:

1. $C_i$: information of every character
2. $E_i$: information of every evidence
3. $T_i$: an array of testimonies
4. $X$ (optional): a context that may provide additional information required for the reasoning

The output of a model is a pair of $(T_i, E_j)$ where an evidence is presented to contradict a testimony.

[5] aceattorney.fandom.com/wiki
[6] lparchive.org/Danganronpa-Trigger-Happy-Havoc/
[7] "Evidence" in Ace Attorney" and "Truth Bullets" in Danganronpa.
[8] "Cross examination" in Ace Attorney and "non-stop debate" in Danganronpa.
[9] Non-core gameplay such as investigation in Ace Attorney or social activities in Danganronpa is lumped into the context.

| Type | Evidence example | Testimony example |
|---|---|---|
| Spatial | Death was caused by a gunshot **to the chest**. | ...fired on the English civilian! And **from the back**... |
| Temporal | Shots were fired **just after midnight** on 12/25. | When she said "**It's almost Christmas**!" shots fired! |
| Causal | ...weapon **bears the defendant's prints**... | I never **touched** the murder weapon. |
| Behavioral | Victim's diary: **Meet with** Hugh. Important. | Huge: I **didn't talk to anyone** until the final bell. |
| Numerical | Cause of death: **single** blunt force trauma. | You see? You hit her **twice**! |
| Physical | The victim was wearing a **plain shirt**. | He was always walking around with a **flowery** shirt. |
| Spelling | The defendant is **Maggey** Byrde. | The blood writing was the defendant's name, "**Maggie**". |

Table 2: Examples (edited for brevity and clarity) of evidences and testimonies of each reasoning type.

At times, there can be multiple ground-truth pairs. Thus, the task is essentially a multiple-choice format with an action space of $|T| \times |E|$, on the order of hundreds. While our dataset is mostly faithful to the original games, we made various types of modification (change of wording, removing turns with loose contradictions, adding information for logic leaps, etc.) to ensure the rigorousness of reasoning.

### 3.2 Annotations

To improve rigorousness of evaluation and enable fine-grained insights into TURNABOUTLLM, we annotate the following aspects of each turn: metadata, reasoning chains, and reasoning types. The annotation protocol is included in C in Appendix.

**Metadata** First, we annotate a one-sentence summary of the current story that provides necessary information for identifying the contradiction for each turn. We provide the span from the evidence and from the testimony that critically constitutes the contradiction. We next label whether a turn is self-contained, where a contradiction can be deducted using only information of characters, evidences, and testimonies, without any other context such as the dialogue transcripts. Whenever a turn is not self-contained, a model needs to perform a needle-in-a-haystack retrieval from the full context (all transcript until the current moment) to gather necessary information (Figure 8). In this case, we manually annotate an expected context span.

**Reasoning Chain** Next, we annotate a reasoning chain used for deriving the contradiction for each turn (Figure 2). A reason chain is a tree structure with three components. First, observed facts, represented as leaf nodes, are paraphrased directly from evidence, testimony, or context. Atomic propositions (non-leaf nodes) are handwritten modus ponens rules that operates upon the facts and derive new facts. Finally, a contradiction (root node) is implied based on two obviously contradiction facts.

As the reasoning in TURNABOUTLLM is based on natural narrative texts, subjectivity in the reasoning chain is unavoidable. Therefore, when annotating the propositions, we uphold the desiderata of only considering general rules in the real world (neglecting what-ifs and extremities) and making them as reasonably atomic as possible.

**Reasoning Types** Lastly, we annotate a fine-grained type of deductive reasoning for each turn. We define 7 reasoning types, including spatial, temporal, causal, behavioral, numerical, physical, and spelling with examples shown in Table 2. We assign one or more types to a turn based on the type of reasoning that underlies the propositions in the annotated reasoning chain (Figure 2). Each reasoning category contains a non-trivial number of turns (Figure 3b), demonstrating that our dataset demands heterogeneous reasoning capabilities.

On average, annotation for each turn takes 20 minutes for a trained annotator, resulting in a total labor of approximately 100 hours.

### 3.3 Statistics

Table 3 summarizes the statistics of TURNABOUTLLM. In total, there are 306 turns in TURNABOUTLLM, with an average of 12 game characters, 38 evidences, 11 testimonies, and 25K text characters.

Figure 3a demonstrates a large answer-space in TURNABOUTLLM, with an average of 200 evidence-testimony pairs to choose from. Figure 3b shows the distribution of different types of reasoning ability required. Combined, these statistics are evidence that TURNABOUTLLM is a challenging and complex benchmark for LLM capabilities.

## 4 Evaluation Protocol

To evaluate a model on the dataset, we extract specific fields from each data point in the game to form a single prompt, and we prompt the model one-time for a single turn. The model is asked to give the indices of the contradicting evidence and testimony. As there may be multiple contradicting pairs in each turn, we regard the output as correct

| Statistics | AA123 | AA456 | GAA12 | AAI12 | DGRP1 | Overall |
|---|---|---|---|---|---|---|
| # Data points | 85 | 72 | 43 | 69 | 37 | 306 |
| Avg. context length (# chars) | 19K | 29K | 36K | 34K | 2.2K | 25K |
| Avg. # characters | 10.6 | 13.6 | 13.2 | 12.6 | 17 | 12.3 |
| Avg./Max. # testimonies | 5.9 / 10 | 5.6 / 8 | 5.7 / 7 | 5.1 / 8 | 6.7 / 11 | 5.7 / 11 |
| Avg./Max. # evidences | 20.2 / 32 | 21.1 / 33 | 18.6 / 30 | 25.3 / 38 | 18.0 / 21 | 21.1 / 38 |
| Avg./Max. length of reasoning chain | 3.5 / 9 | 3.8 / 10 | 3.6 / 6 | 3.5 / 8 | 3.3 / 5 | 3.6 / 10 |

Table 3: Overall statistics of TurnaboutLLM, categorized by the incorporated detective game titles. **AA123** stands for *Phoenix Wright: Ace Attorney Trilogy*. **AA456** stands for *Apollo Justice Ace Attorney Trilogy*. **GAA12** stands for *The Great Ace Attorney Chronicles*. **AAI12** stands for *Ace Attorney Investigations Collection*. **DGRP1** stands for *Danganronpa: Trigger Happy Havoc*.

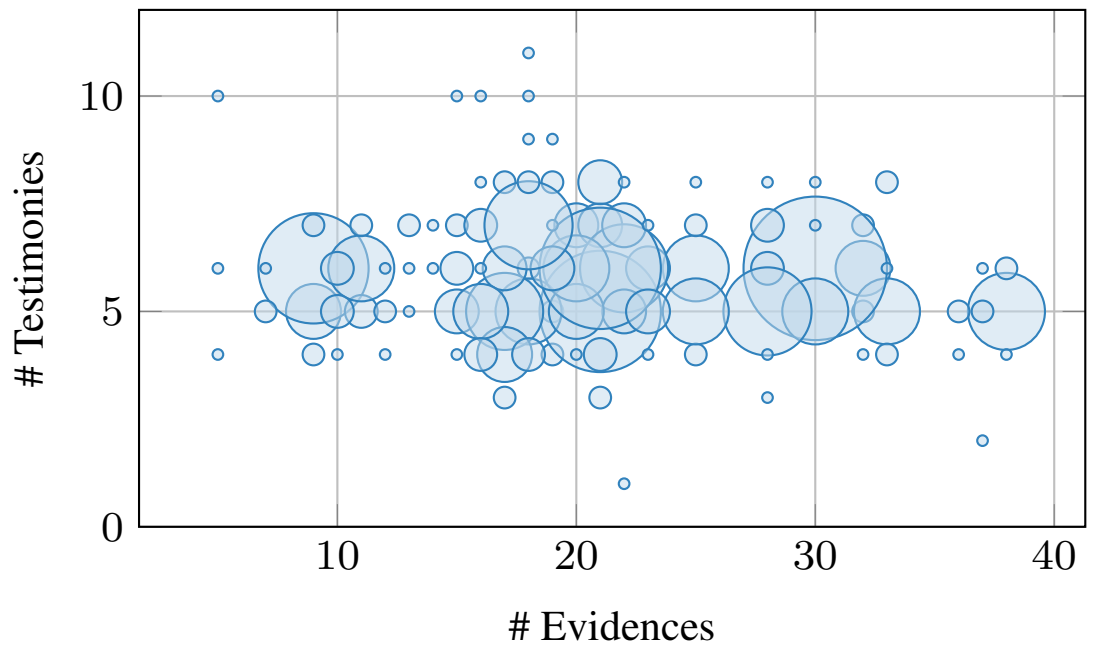

(a) An illustration of the number of turns in TurnaboutLLM (size of each circle) with respect to the number of available evidences (horizontal) and testimonies (vertical) to choose from.

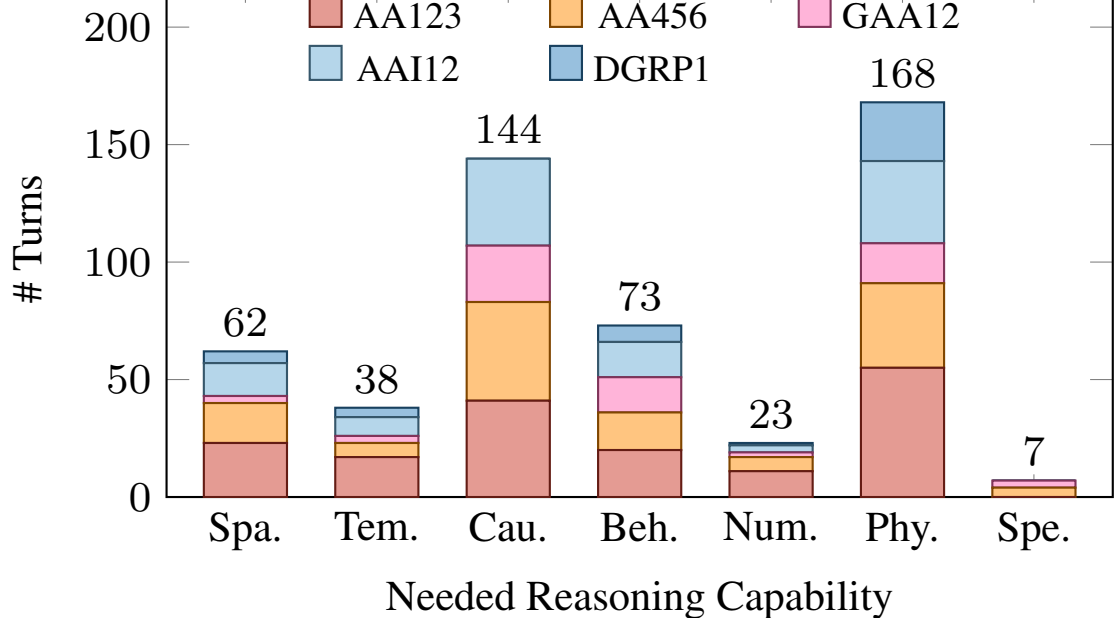

(b) The number of TurnaboutLLM turns with respect to the reasoning capabilities required (e.g., Spatial, Temporal, etc.) to find the contradiction, classified by the incorporated title.

Figure 3: Illustrations of further statistics of our TurnaboutLLM dataset.

if the proposed pair is included in the list of ground truth contradicting pairs.

**Evaluation Metrics** We compute the overall accuracy of the model as the percentage of correct answers across all turns, and we compute the evidence accuracy and testimony accuracy respectively as the percentage of correct evidence and testimony presented across all turns.

**Data Splits** We do not endorse any particular train-develop-test split of TurnaboutLLM and leave that decision to future users. In this work, we treat the entirety of the Ace Attorney dataset as the evaluation set, since we do not attempt any hyperparameter tuning or modeling improvement.

**Evaluation Settings** To better gauge different aspects of models' reasoning abilities, we propose 4 variations of the evaluation prompt templates based on available property fields in the data. First, We start with a **basic** zero-shot prompt[10] with an average of 1,686 words, which sequentially includes descriptions of all the characters, evidences, and testimonies in the current turn. In case more context than mere evidence descriptions are needed for reasoning, we append a short "context span", an excerpt from the context field that guarantees to fills in the most relevant context information, to the corresponding evidence description.

Second, we use a one-shot, **Chain-of-Thought (CoT)** prompt with an average of 2,280 words, which contains an one-shot example to demonstrate how to reason through evidences to find the contradiction. Moreover, we append a "let's think step by step" instruction to elicit reasoning behaviors. We do this for all non-reasoning models excluding DeepSeek-R1 and OpenAI's o-series. We also provide further ablations on individual effects of one-shot and CoT prompting in Appendix D.

Third, we use a **full-context** prompt averaging 44K words, which includes the complete context of all prior turns within the same court case leading up to the current one. This is a challenging but realistic setting, as all human players experience the game this way. As such, needle-in-a-haystack retrieval of critical information from the context is necessary for turns that are not self-contained by merely characters, evidences, and testimonies.

Fourth, to study whether the model is memorizing the game from its training corpus, we provide an **ablation** prompt with an average of 537 words

[10] Our experiments show that few-shot prompting leads to worse results which are omitted.

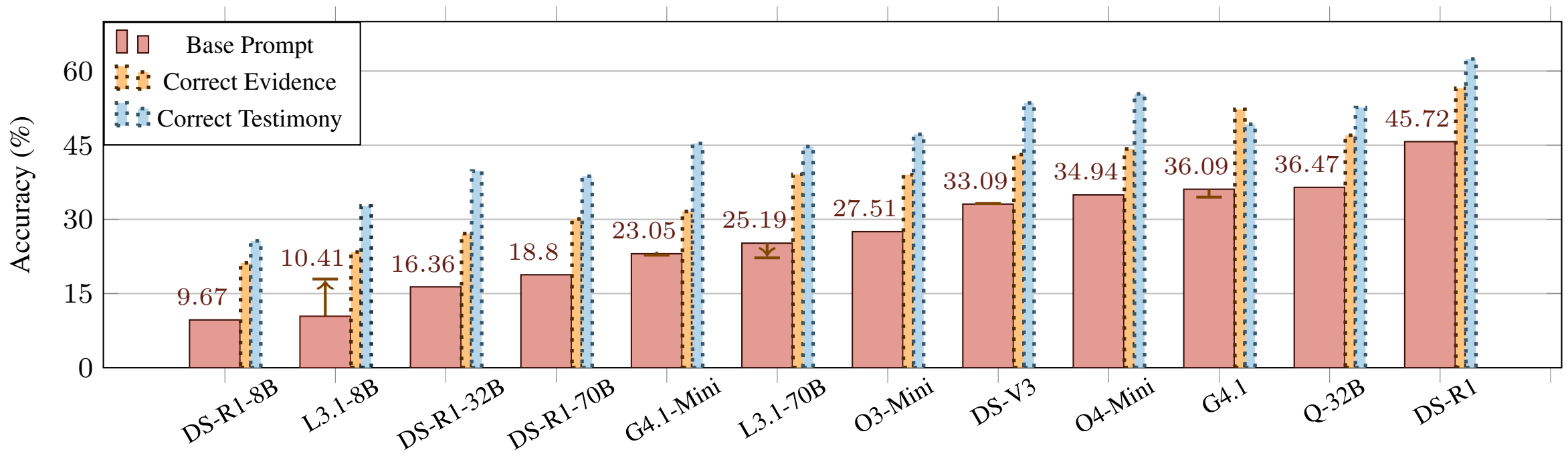

Figure 4: Performance comparison on TURNABOUTLLM across 12 models, ordered from left to right. Bars indicate correctness accuracy (%) using a base prompt, along with accuracy for evidence and testimony. For models without native reasoning capabilities, arrows show the performance change when applying chain-of-thought prompting.

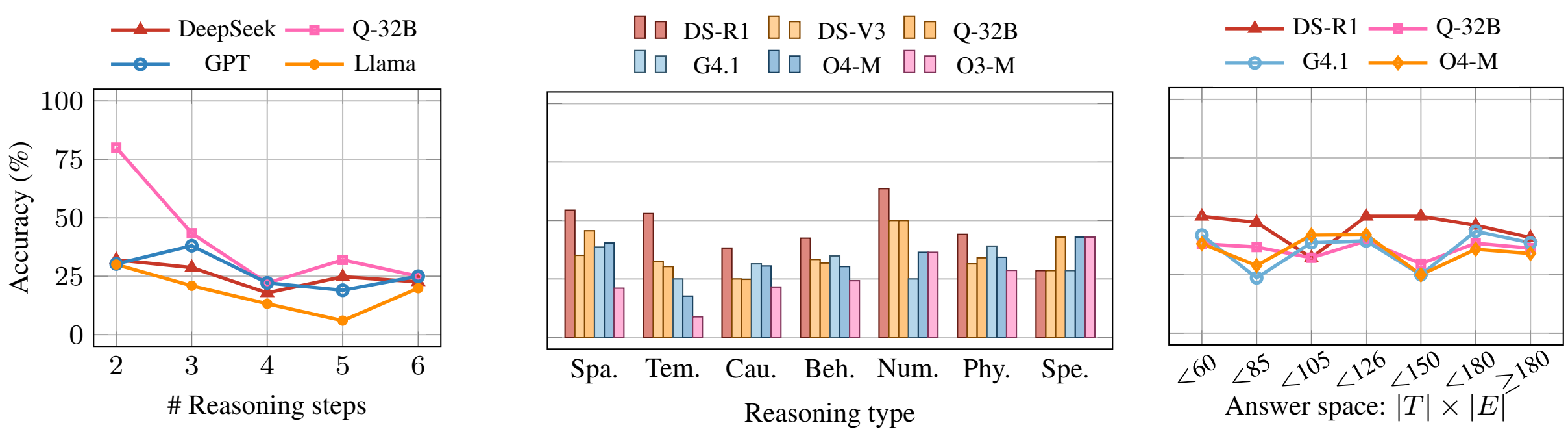

(a) Average accuracy among each model family declines as the number of annotated reasoning steps increases.

(b) Accuracy with respect to the reasoning types. While performance vary a lot across models, causal reasoning is usually the weakest.

(c) Accuracy with respect to size of answer space. Results does not show strong negative correlation.

Figure 5: Model accuracies plotted against the number of reasoning steps, required reasoning types, and size of answer space. Due to space constraints, we only show the performance of 6 representative models. A more comprehensive illustration is shown in the appendix.

where all descriptions of the characters and evidences are removed. The model will have to reason based on the names of the characters and evidences alone, which is often insufficient. Therefore, we would expect a significant drop in its performance if it does not memorize key events in the game.

As is previously discussed, evidences and sometimes testimonies come with images that are occasionally crucial for reasoning about the contradiction. While we have fully captioned them in this work, we also provide all the images and clearly label whenever they are required so that a multimodal evaluation is available for future work.

**Experiments** We evaluate 12 LLMs on our 4 variations of prompts. The LLMs come from 4 model families: the DeepSeek series which includes the 671B DeepSeek-R1 (DS-R1) and V3 (DS-V3) and the smaller distilled DeepSeek-R1-70B (DS-R1-70B), DeepSeek-R1-32B (DS-R1-32B), and DeepSeek-R1-8B (DS-R1-8B) models, the OpenAI family including GPT-4.1 (G4.1), GPT-4.1-mini (G4.1-M) and the reasoning models o3-mini (O3-M) and o4-mini (O4-M), the Llama-3.1-instruct family including Llama-70B (L3.1-70B) and Llama-8B (L3.1-8B), and the reasoning model QwQ-32B (Q-32B) exceling in reasoning and coding. Except for OpenAI models and the two largest DeepSeek models that are run via their APIs, we run all other models locally on 8 H100 GPUs using HuggingFace and KANI (Zhu et al., 2023) .

## 5 Results and Analysis

In this section, we present our primary empirical findings regarding LLMs' reasoning abilities. We begin by highlighting the overall accuracies of all 12 models on TURNABOUTLLM summarized in Figure 4. Subsequently, we provided detailed analyses that dissect model performance by factors such as numbers of reasoning steps (Figure 5a), reasoning types (Figure 5b), answer space sizes (Figure 5c), numbers of reasoning tokens (Figure 6) and prompting strategies (Figure 4, 7).

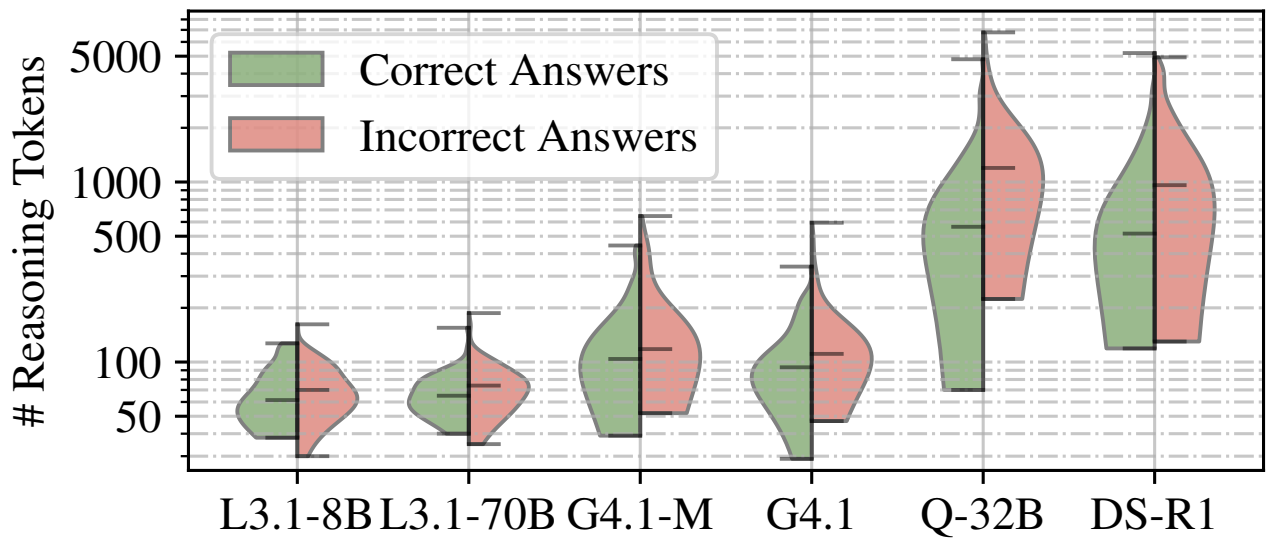

Figure 6: Distributions of the number of generated reasoning tokens, separated by whether a correct answer is derived.

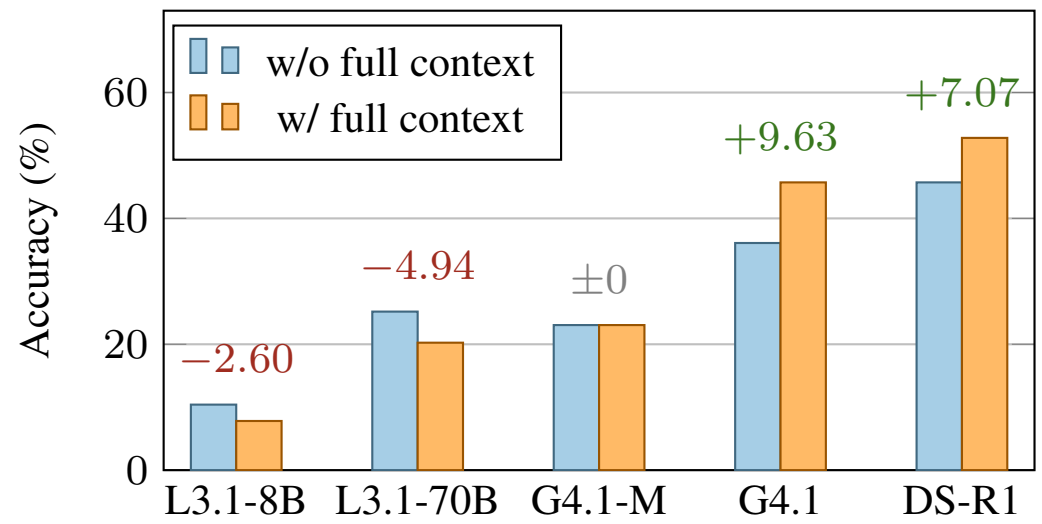

Figure 7: Model performance with or without providing full story context within the prompt.

**The dataset poses a significant challenge in long-context deductive reasoning for state-of-the-art models.** All 12 models demonstrate considerable diffuculty in correctly identifying evidence-testimony pairs within TURNABOUTLLM (Figure 4). Among them, DS-R1 achieves the highest accuracy of 45.72% using the basic prompt. All models, except G4.1, achieve higher accuracy in selecting the correct evidence than in selecting the correct testimony. This trend aligns with the fact that there are typically fewer candidate evidences than testimonies to evaluate. These findings illustrate that TURNABOUTLLM represents a substantial challenge for even the most advanced LLMs.

**Minimal memorization makes the dataset a reliable benchmark for LLMs.** Evaluating 4 models on the ablation prompt with no evidence descriptions, we consistently observe scores at a merely average of 15%, suggesting little memorization. On the contrary, the models' reasoning traces reveal that they are making the most likely "bet" based on evidence names alone. For example, from the evidence name "Poison Gas Ingredients", R1 infers that the item likely contradicts a testimony about the poison gas containing Normallium, which happens to be the correct contradiction. Therefore, we conclude that major models only have minimum memorization of TURNABOUTLLM, which establishes it as a fair ground for LLM evaluations.

**Incorrect results consume more reasoning tokens than correct ones, and more output tokens do not necessarily yield better results.** We define "reasoning tokens" as intermediate tokens generated by the model before arriving at the final answer. Across all models, incorrect responses exhibit higher median and maximum numbers of reasoning tokens compared to correct ones (Figure 6), indicating a negative correlation between model accuracy and the number of reasoning tokens. This potentially shows that when the model produces incorrect answers, outputing additional reasoning tokens does not yield more improvements.

We observe a surplus of reasoning tokens produced by Q-32B and DS-R1 over other models in Figure 6 using a logarithmic scale. However, *despite using far fewer reasoning tokens than Q-32B, G4.1 achieves approximately equal accuracy, exhibiting superior reasoning efficiency under a limited token budget.* This could further corroborate with the conjecture that intentional exploration of the answer space is more decisive to model performance than extensive output of reasoning tokens.

**Full context benefits large models but hurts smaller ones.** Including the complete context in the evaluation prompt has contrasting effects depending on the size of the model (Figure 7). Large models such as G4.1 and DS-R1 exhibit notable accuracy improvements of approximately 15% compared to their basic prompt performances. Conversely, small and medium-sized models, such as L3.1-70B and L3.1-8B, suffer performance declines. This could suggest that smaller models, limited by their parameter size, not only under-utilize additional contextual information but are also "confused" by the influx of supplementary data.

**Models struggles the most with extracting key facts in deductive reasoning.** Sampling 5% of all incorrect responses from DS-R1, Q-32B, and G-4.1, we categorize reasoning errors into five categories: (i) extracting false factual information; (ii) selecting the wrong fact for deduction; (iii) generating the wrong proposition given correct facts; (iv) generating degenerate deductions given correct premises; and (v) prioritizing less rigorous deductions. As Table 5 shows, (i) is the most common mode of failure, suggesting that models predominantly have more difficulty understanding critical context than following deductive steps.

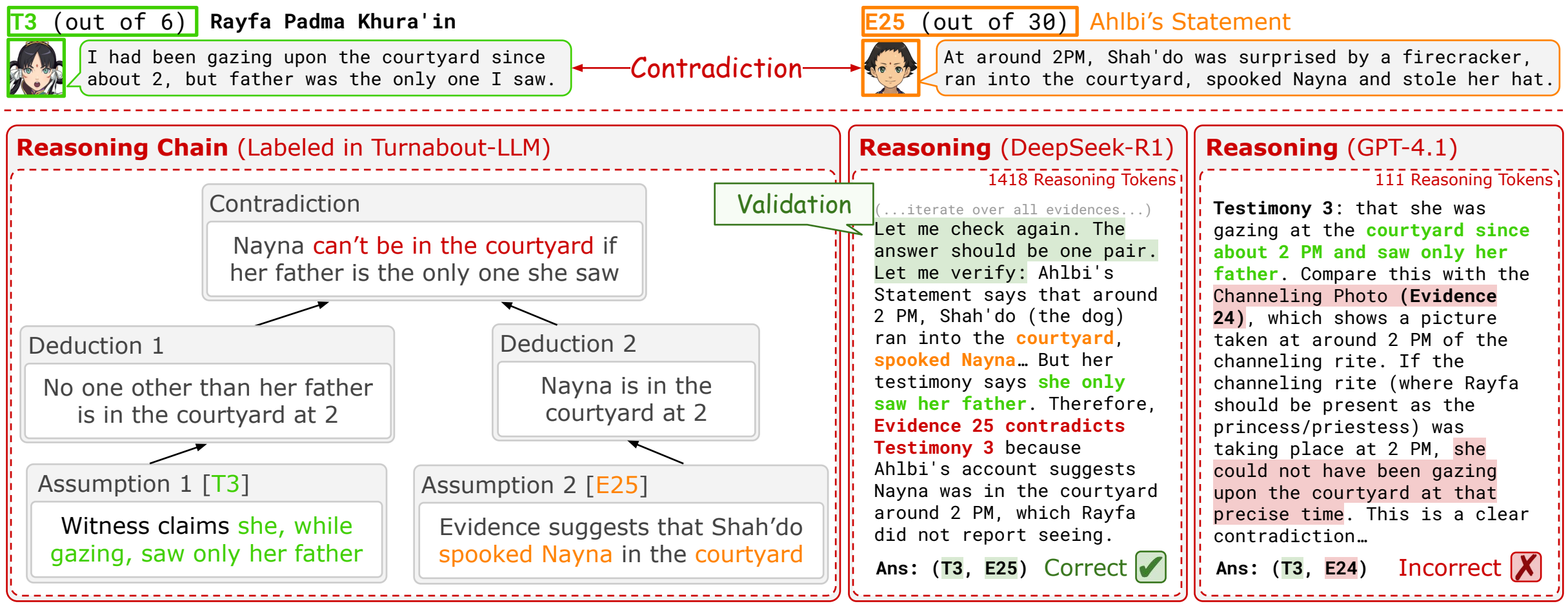

Figure 8: A qualitative comparison between DeepSeek-R1 and GPT-4.1's reasoning on answering the 2nd turn of AA6-5-4. GPT-4.1 failed by jumping straight into conclusion, while DS-R1 carefully examines all evidences and testimonies, producing over 1.4K reasoning tokens as well as the correct answer.

**Model performance deteriorates with increasing reasoning steps, but not with larger answer spaces.** There is a negative correlation between average accuracy within a model architecture family and the number of reasoning steps (Figure 5a). As the number of reasoning steps increases, performance gradually declines, signaling that questions requiring more logical connections tend to be more difficult. This supports the validity of using annotated reasoning chains as an indicator of difficulty.

In contrast, the size of the answer space does not appear to impact model accuracy (Figure 5c). By categorizing answer spaces into seven bins with approximately equal numbers of data points, we observe consistent model performance across all bins. Further analysis reveals that reasoning models tend to use many reasoning tokens to exhaustively enumerate possible testimony-evidence pairs without engaging in deeper reasoning.

**CoT prompting does not enhance model performance.** We notice minimal benefits of CoT prompting on reasoning performance (see Figure 4). For all 5 models except the smallest L3.1-8b, this prompting method either results in no improvement or minor performance decreases. The models' reasoning traces reveal that CoT prompting delays the time the model first reaches its final conclusion and allows it to "think" more. However, the extended thinking often hinges on a single evidence-testimony pair, failing to conduct an extensive search in the answer space. This appears to imply that CoT prompting is ineffective in solving deductive reasoning tasks with extensive answer spaces and large context sizes.

**Models benefit from longer explorations of the answer space.** Models can effectively extend explorations of the answer space to boost their accuracy, as is shown by the qualitative example in Figure 8. In the example, we observe distinct behaviors in G4.1 and DS-R1's reasoning traces. G4.1, generating only 111 tokens, merely considers one possible evidence before finalizing on a wrong answer. In contrast, DS-R1, generating 1,418 tokens, explores multiple evidences before narrowing down to 3 most likely candidates and arriving at the correct answer. We conjecture that when in a large answer space, successful deductive reasoning is grounded in extensive, trial-and-error search and does not have a cognitive shortcut.

**Different models excel at different reasoning types and scale with increasing parameter size.** Different models have particular strengths and weaknesses depending on the type of reasoning required (Figure 5b). Models generally perform best on numerical tasks involving counting and comparison, whereas most exhibit their lowest scores on temporal or causal reasoning. Furthermore, model performance tends to improve as the parameter size increases (Figure 4), with the notable exception of Q-32B, which outperforms all larger models except the 671B DS-R1. The positive correlation between parameter size and model accuracy could imply that larger models may possess inherently stronger deductive reasoning capabilities.

## 6 Conclusion

We introduce TURNABOUTLLM , the first benchmark that embeds symbolic-logic puzzles inside narrative-rich, super-long contexts drawn from detective visual novels. By performing an extensive empirical study across twelve contemporary LLMs, we show that TURNABOUTLLM is challenging and poses a fair ground to evaluate LLMs' reasoning abilities. We release the dataset, annotation toolkit, and evaluation code to spur research on (i) scalable long-context reasoning, (ii) controllable chain-of-thought generation, and (iii) unified metrics for symbolic-narrative tasks. We hope TURNABOUTLLM will serve as a stepping-stone toward LLMs that can navigate the messy, open-world logic of real human discourse.

## 7 Limitation

Despite its breadth, TURNABOUTLLM still faces several constraints. First, its detective-courtroom focus targets contradiction spotting, leaving other deductive settings—such as scientific discovery or regulatory compliance—largely untested. Second, because the narratives originate from Japanese visual novels, they may encode culture-specific norms and idioms that bias evaluation toward models already familiar with such text. Third, although we supply descriptive captions for in-game images, true multimodal reasoning is only approximated, not fully exercised. Fourth, the dataset's manually crafted reasoning chains ($\approx$ 100 annotator-hours) introduce subjectivity and hamper scalability, though future releases will report inter-annotator agreement and provide semi-automated validation tools. Fifth, while the raw scripts are publicly available, their copyright status could change; We are committed to honoring any takedown requests from the rights holders. Finally, evaluation with 100K-token prompts imposes a heavy computational footprint, and researchers with limited resources may need chunk-wise retrieval strategies that we have not yet benchmarked. Acknowledging these limitations helps define the benchmark's current scope and highlights directions for future expansion.

## Acknowledgment

We thank Sesh Sadasivam for the initial ideation of this work. We thank Manvi Kaul for the initial efforts of modeling. We thank Bowen Jiang for her wonderful comments on and edits to the writing of this paper. We thank Shu Takumi, Kazutaka Kodaka, and their teams for the marvelous gift to the Ace Attorney and Danganronpa community that makes this work possible.

## A License and Intended Use

The data utilized in this research is sourced from fandom.com. As stipulated by fandom.com, their resources are made available under the Creative Commons Attribution-Share Alike License 3.0 (Unported) (CC BY-SA). This license permits the sharing and adaptation of the material, provided that appropriate attribution is given to the original source, a link to the license is provided, and that if the material is remixed, transformed, or built upon, the contributions are distributed under the same or a compatible license. Our intended use of this data is strictly for academic research and analysis within this paper, fully adhering to the terms and conditions set forth by the CC BY-SA license.

## B Annotator demographics

Five annotators contribute to authoring and verifying each data point's reasoning types, reasoning steps, and evidence and context span. All are U.S.-based university students and avid Ace Attorney and Danganropa players, thus ideally suited to examine each case data's key attributes.

# C Annotation Protocol

The two primary assets that are manually annotated are **labels of reasoning types** and **semi-formal reasoning chains** for each turn. The annotation is performed in two stages by four annotators in total, following the guidelines below.

In the first stage, all annotators *individually* perform and then *discuss* the annotations on the first two installments of the game (AA1-2) iteratively until they reach **100% inter-annotator agreement**. During this process, the annotation guideline is refined and finalized.

In the second stage, two out of the four annotators each annotate half of the remaining turns (in AA3-6, Danganronpa1) individually. Their annotations are then validated by a third annotator, where a handful of disagreements are resolved. No further cross-validation is performed due to the prohibitively high cost of annotation.

## C.1 Labels

**Numerical.** Labeled as "numerical" only if the core contradiction is a difference in numbers. Example: "I heard 1 gunshot" vs. "2 gunshots were fired". Numbers that do not constitute a contradiction should **not** be labeled as numerical. Example: "I gave the victim 2 items" vs. "The witness never met the victim".

**Temporal.** Labeled as "temporal" only if the core contradiction involves time. Example: "He died before noon" vs. "Time of death is after 3 PM". This may stack with other labels like "numerical". Mentions of time that do not constitute a contradiction should **not** be labeled temporal. Example: "I met the victim in the morning" vs. "The witness never met the victim".

**Spatial.** Labeled as "spatial" only if the core contradiction involves space. Example: "I killed him at the bus stop" vs. "The victim was found dead in his home". Mentions of location without contradiction should **not** be labeled spatial. Example: "I met the victim at school" vs. "The witness never met the victim".

**Physical.** Labeled as "physical" only if the core contradiction involves non-universal physical properties of an object. Example: "I saw him beaten by a club" vs. "Autopsy report shows only trauma of piercing". This object cannot be time or space but can be an abstract concept. Example: "I never told anyone this idea" vs. "The victim wrote down this idea in a notebook". If the contradiction involves human behavior, it should be labeled as "behavioral" instead. At times, this may stack with other labels such as temporal or spatial. Example: "I did not hear anything from the clock at noon" vs. "The clock sounds at noon" (also temporal) "I saw the vase" vs. "There was a wall between the witness and the vase" (also spatial)

**Behavioral.** Labeled as "behavioral" only if the core contradiction involves human behavior, such as intent, habits, or preferences. Example: "Larry hates music" vs. "Larry is reported to listen to music every day". Exceptions are only considered if there is strong evidence. If another contradiction type applies, the label should not be behavioral. Example: "I killed him at the bus stop" vs. "The victim was found dead in his home" may lead to a corollary of "I killed him" vs. "I cannot have killed him".

**Spelling.** Labeled when the contradiction is due to spelling differences. Example: "Harry" vs. "Henry".

## C.2 Reasoning Chain

A **reasoning chain** is a manually annotated list of **facts** or **propositions** that lead to a contradiction.

**Considerations:**

- A **fact** is a paraphrase of the testimony span ("I saw the victim getting shot."), evidence span ("only piercing wounds were found."), or context span.
- A **proposition** is a general rule of implication. Example: "if someone gets shot, there will be ballistic wounds, not piercing wounds."
- Entities in propositions should be lifted and generalized when possible. Example: use "someone" in lieu of "Mr. Tanaka".
- There must be at least one proposition. Propositions are framed as *Modus ponens*:

  $$\text{Assertion P} + \text{Conditional} \Rightarrow \text{Assertion Q}$$

- Each fact and proposition should be as atomic as possible, though some subjectivity is inevitable.

| Model | Prompt | Examples | Overall (%) | Testimony (%) | Evidence (%) |
|---|---|---|---|---|---|
| DS-V3 | Base | Zero-shot | 33.09 | 53.53 | 43.12 |
| | Base | One-shot | 31.23 | 55.02 | 41.26 |
| | CoT | Zero-shot | 40.52 | 57.25 | 52.79 |
| | CoT | One-shot | 33.46 | 50.19 | 44.61 |
| G-4.1 | Base | Zero-shot | 36.09 | 49.25 | 52.26 |
| | Base | One-shot | 37.55 | 57.25 | 47.96 |
| | CoT | Zero-shot | 40.15 | 56.88 | 56.13 |
| | CoT | One-shot | 34.20 | 51.30 | 45.72 |
| L3.1-70B | Base | Zero-shot | 25.19 | 44.74 | 39.10 |
| | Base | One-shot | 7.81 | 29.00 | 15.24 |
| | CoT | Zero-shot | 10.27 | 23.77 | 18.11 |
| | CoT | One-shot | 21.93 | 40.15 | 34.20 |

Table 4: Performances of four setups that separate the effects of one-shot vs CoT prompting across three models.

| Model | Fact extraction (%) | Fact selection (%) | Proposition generation (%) | Degenerate deduction (%) | Deduction ranking (%) |
|---|---|---|---|---|---|
| DS-R1 | 47.1 | 17.7 | 23.5 | 17.7 | 29.4 |
| Q-32B | 44.4 | 0.0 | 22.2 | 0.0 | 33.3 |
| G-4.1 | 45.5 | 27.3 | 18.2 | 0.0 | 0.0 |

Table 5: Common error types made by each model when their answers were incorrect. The percentages indicate the proportion of incorrect answers falling into each error category.

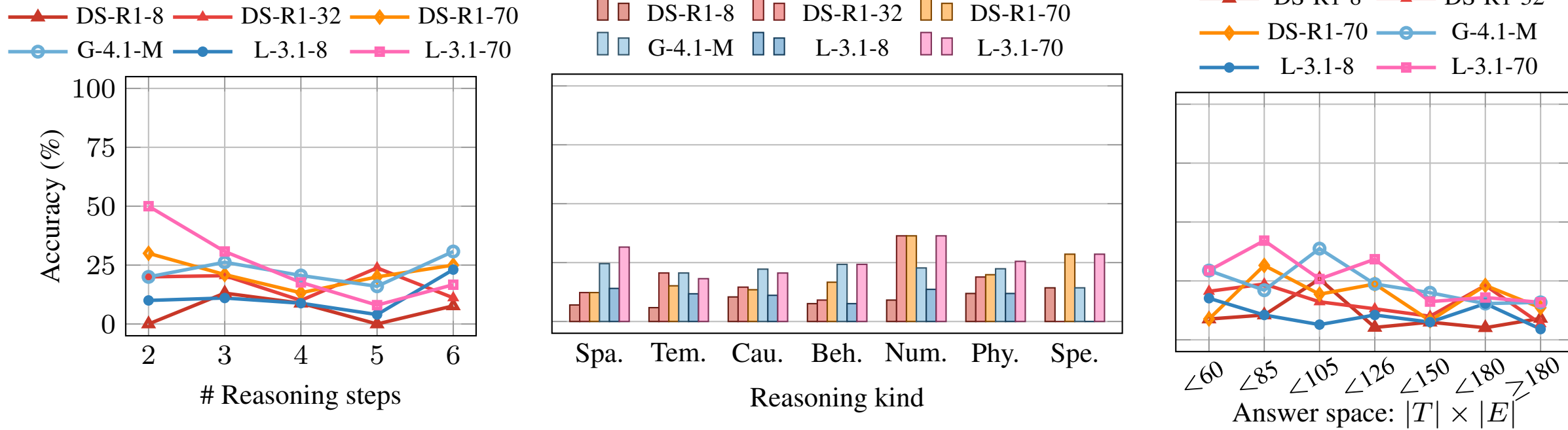

(a) Accuracy decreases as the number of reasoning steps grows. Due to scarcity, we omit problems that need $> 6$ steps.

(b) Accuracy with respect to the reasoning types. While performance vary a lot across models, temporal reasoning is usually the weakest.

(c) Accuracy with respect to size of answer space. Results does not show strong negative correlation.

Figure 9: Model accuracies plotted against the number of reasoning steps, required reasoning types, and size of answer space. Additional experiments not covered in the main body text are presented here.

## D Ablations on one-shot prompting and CoT prompting

To decouple the effects of one-shot vs CoT prompting in our pipeline, we design four prompt variations that carefully control the target variable. They are: (i) zero-shot prompting; (ii) one-shot prompting with an example; (iii) zero-shot CoT prompting; and (iv) one-shot prompting with an example. We conduct the four experiments on three non-reasoning models, DS-V3, G-4.1, and L3.1-70B, on the whole TURNABOUTLLM dataset.

As shown in Table 4, we observe that: (i) zero-shot CoT prompting greatly helps large models (DS-V3 and G-4.1), where both attain their best performance out of four, but it reduces L3.1-70B's accuracy by more than half. (ii) one-shot example offers little help to the two large models on both base and CoT prompting, but it has opposite effects when applied to L3.1-70B in both scenarios.

## E Additional Data Examples and Statistics

Figure 10 and 11 present two highly challenging examples from TURNABOUTLLM. Figure 9 shows additional performance breakdown of models that are not included in the main section.

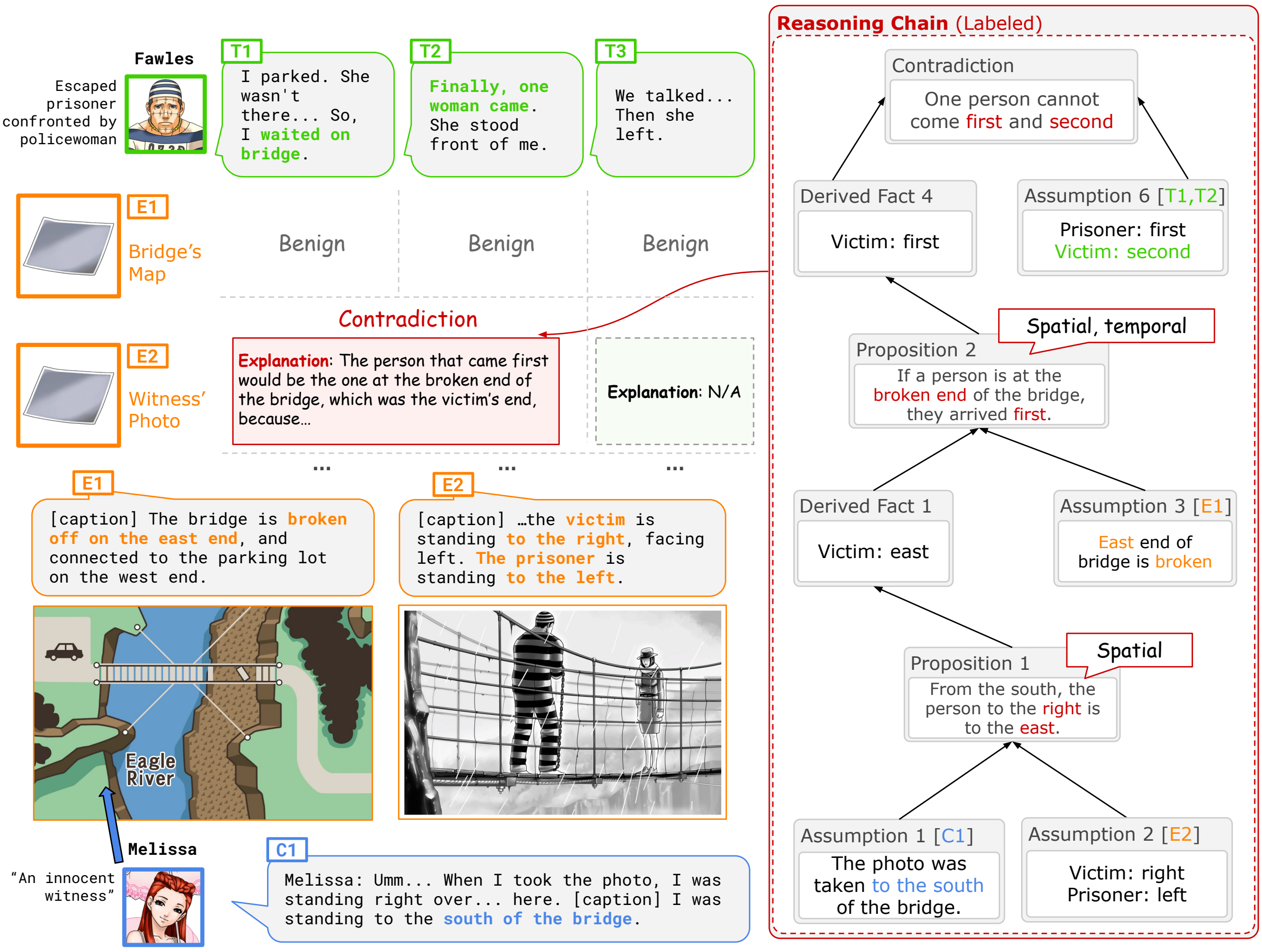

Figure 10: A highly challenging data point from TURNABOUTLLM involving spatial and temporal reasoning.

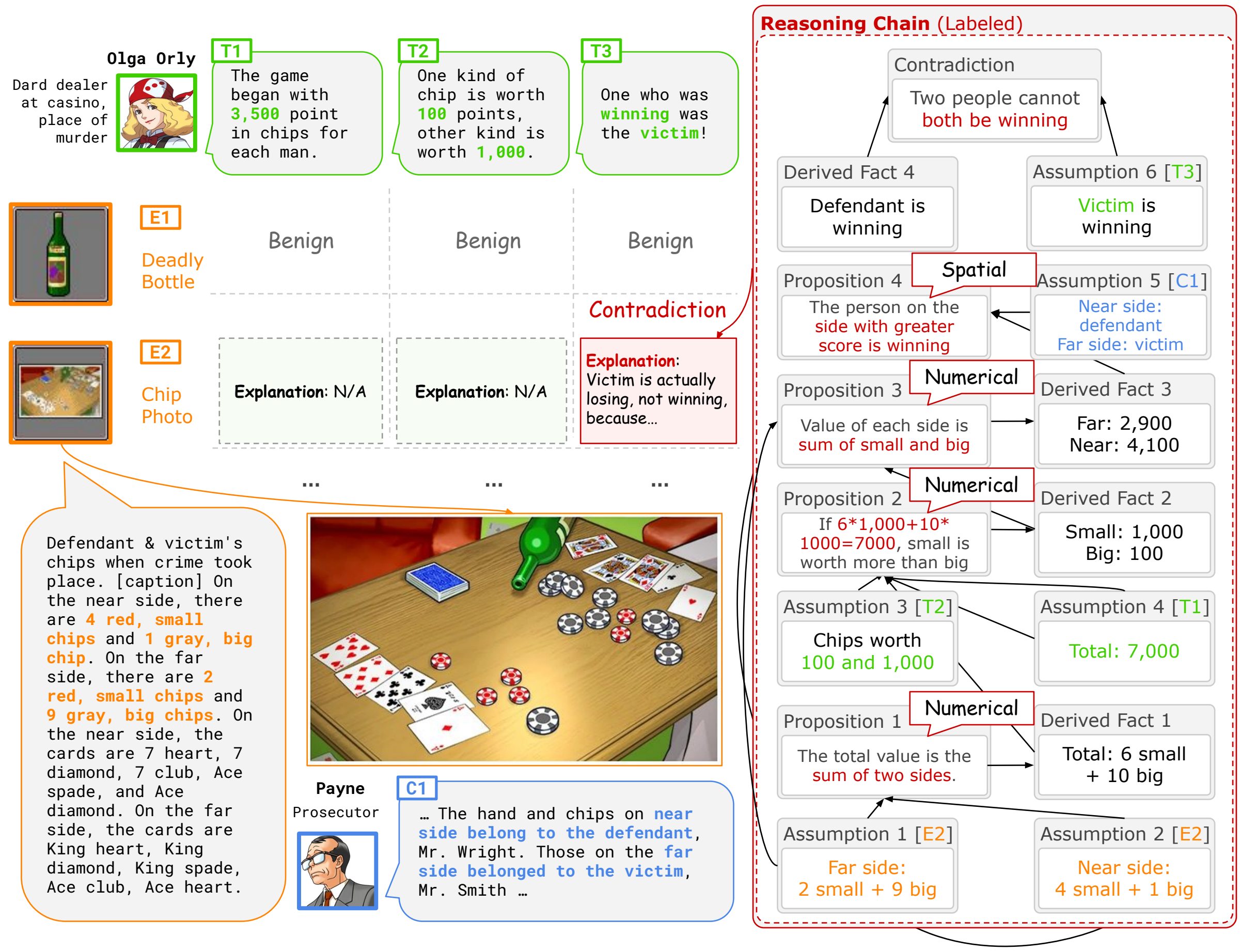

Figure 11: A highly challenging data point from TURNABOUTLLM involving numerical and spatial reasoning, even with a touch of abductive reasoning.